\pdfoutput=1
\documentclass[letterpaper, 10 pt, conference]{ieeeconf}  
\IEEEoverridecommandlockouts                              
\usepackage{hyperref}
\usepackage{graphicx}
\usepackage{cleveref}
\usepackage[utf8]{inputenc}
\usepackage[T1]{fontenc}
\usepackage[dvipsnames]{xcolor}
\usepackage{dirtree}
\usepackage{makecell}
\usepackage{graphicx}
\usepackage{subcaption}
\usepackage[english]{babel}
\usepackage{blindtext}
\usepackage{cite}
\usepackage{kotex}
\usepackage{caption}
\usepackage{multicol}
\usepackage{multirow}
\usepackage{array}
\usepackage{arydshln}
\usepackage{amssymb}
\usepackage{pifont}
\usepackage[switch,columnwise]{lineno}

\title{\LARGE \bf
Run Your Visual-Inertial Odometry on NVIDIA Jetson\\: Benchmark Tests on a Micro Aerial Vehicle}

\author{Jinwoo Jeon$^{\dagger}$, Sungwook Jung$^{\dagger}$, Eungchang Lee, Duckyu Choi,~\IEEEmembership{Student~Member,~IEEE}, \\and Hyun Myung$^{*}$,~\IEEEmembership{Senior~Member,~IEEE}
\thanks{$^{\dagger}$These authors contributed equally to this work.}
\thanks{*This work was partially supported by BK21 FOUR and Korea Ministry of Land, Infrastructure and Transport 362 (MOLIT) as `Innovative Talent Education Program for Smart City'.}
\thanks{$^{\dagger}$J. Jeon, E. Lee, D. Choi, and $^{*}$H. Myung are with Urban Robotics Laboratory, 
        Korea Advanced Institute of Science and Technology, Daejeon, Republic of Korea
        {\tt\small \{zinuok, sungwook87, eungchang\_mason, duckyu, hmyung\}@kaist.ac.kr}}%
\thanks{$^{\dagger}$S. Jung is with Autonomous IoT Research Center, Korea Electronics Technology Institute (KETI), Seongnam 13509, Republic of Korea
        {\tt\small \{sungwook87\}@keti.re.kr}}%
\thanks{$^{*}$H. Myung is also with KI-AI, and KIR at KAIST}
}

\begin{document}

\maketitle
\thispagestyle{empty}
\pagestyle{empty}
\begin{abstract}
This paper presents benchmark tests of various visual(-inertial) odometry algorithms on NVIDIA Jetson platforms. The compared algorithms include mono and stereo, covering Visual Odometry (VO) and Visual-Inertial Odometry (VIO): VINS-Mono, VINS-Fusion, Kimera, ALVIO, Stereo-MSCKF, ORB-SLAM2 stereo, and ROVIO. As these methods are mainly used for unmanned aerial vehicles (UAVs), they must perform well in situations where the {size of the processing board} and weight is limited. Jetson boards released by NVIDIA satisfy these constraints as they have a sufficiently powerful central processing unit (CPU) and graphics processing unit (GPU) for image processing. However, in existing studies, {the performance of Jetson boards} as a processing platform for executing VO/VIO has not been compared extensively in terms of the usage of computing resources and accuracy. Therefore, this study compares representative VO/VIO algorithms on several NVIDIA Jetson platforms{, namely} NVIDIA Jetson TX2, Xavier NX, and AGX Xavier, and introduces a novel dataset 'KAIST VIO dataset' for UAVs. Including pure rotations, the dataset has several geometric trajectories that are harsh to visual(-inertial) state estimation. The evaluation is performed in terms of the accuracy of estimated odometry, CPU usage, and memory usage on various Jetson boards, algorithms, and trajectories. We present the {results of the} comprehensive benchmark test and release the dataset for the computer vision and robotics {applications.}
\end{abstract}

\section{INTRODUCTION}


Recently, unmanned aerial vehicles (UAVs) {have become increasingly} important and applicable in various fields, including structural inspection\cite{jung2018multi,jung2020bridge}, environment monitoring\cite{jung2017development,kim2016image}, and surveillance\cite{scherer2020multi}. {It is crucial that a UAV should be able to estimate its state accurately in real time for an autonomous flight system.} Therefore, there has been a massive effort to develop precise state estimation algorithms. However, the UAV system has limitations in terms of size, payload, and power, {which are problems that are commonly encountered in the filed of computer vision and robotics.}


Visual odometry (VO) has been solving these issues using vision sensors. The most widely used vision sensor for the VO method is a monocular camera. {Unlike other sensors, this sensor is economical, compact, and power efficient. Hence, they can be easily mounted on a UAV.} However, it is impossible to obtain the absolute scale of the traveled path using only monocular images captured by the camera. In the field of computer vision and robotics, this scaling problem has been solved in various ways. The RGB-D sensor\cite{rgbd,whelan2013robust}, deep learning-based methods\cite{deep-depth-1,deep-depth-2}, and stereo vision\cite{stereo,gomez2016robust} have been used to obtain the depth information to infer the absolute scale. {Another commonly used approach} is to combine additional sensors with the camera to obtain additional information for measuring the movement of the camera attached to the rigid body of the robot.
\begin{figure}[t]
\begin{subfigure}{.25\textwidth}
  \centering
  \includegraphics[width=4.2cm]{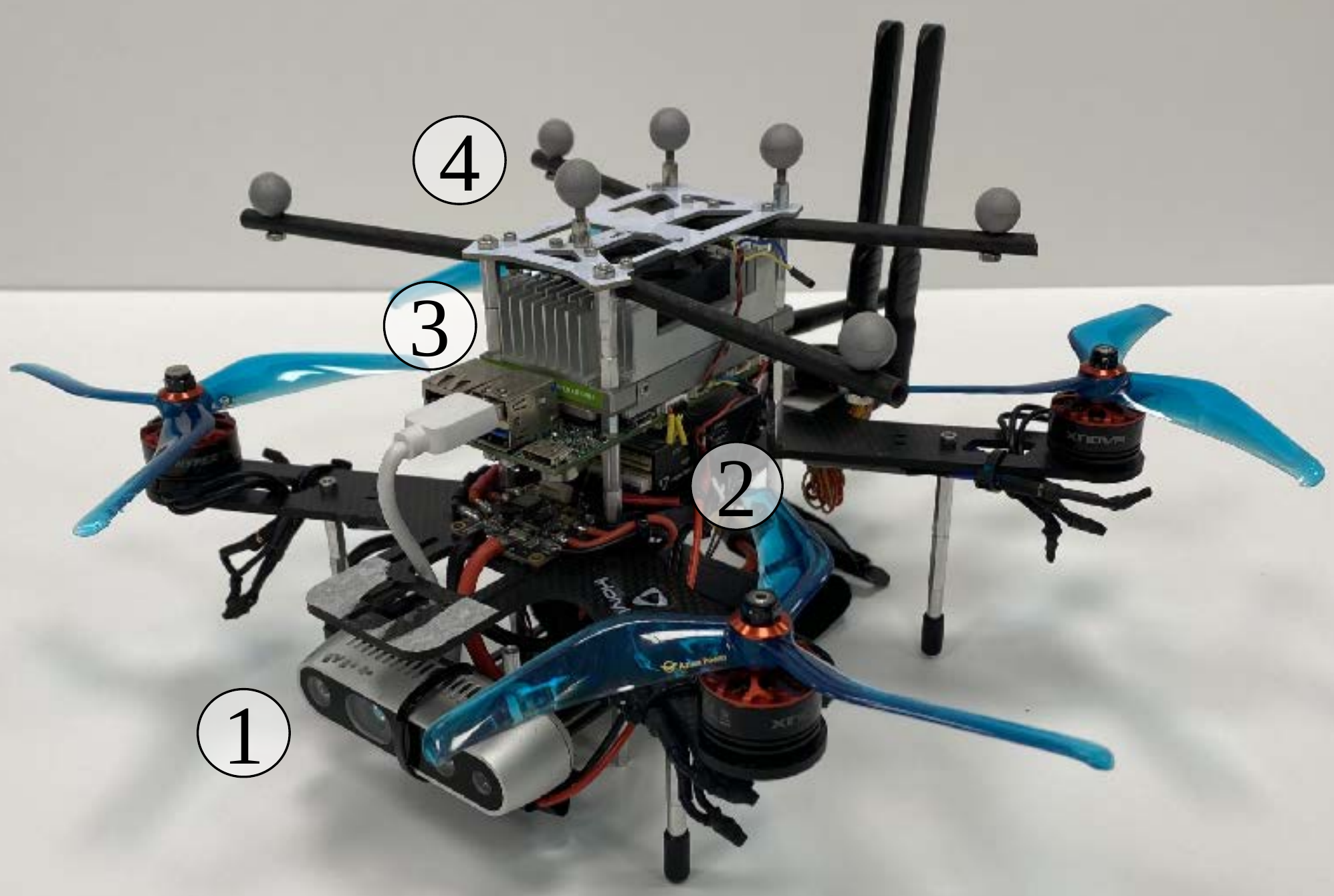}\hfill%
  \caption{}
  \label{fig:uav}
\end{subfigure}
\begin{subfigure}{.23\textwidth}
  \centering
  \includegraphics[width=\textwidth]{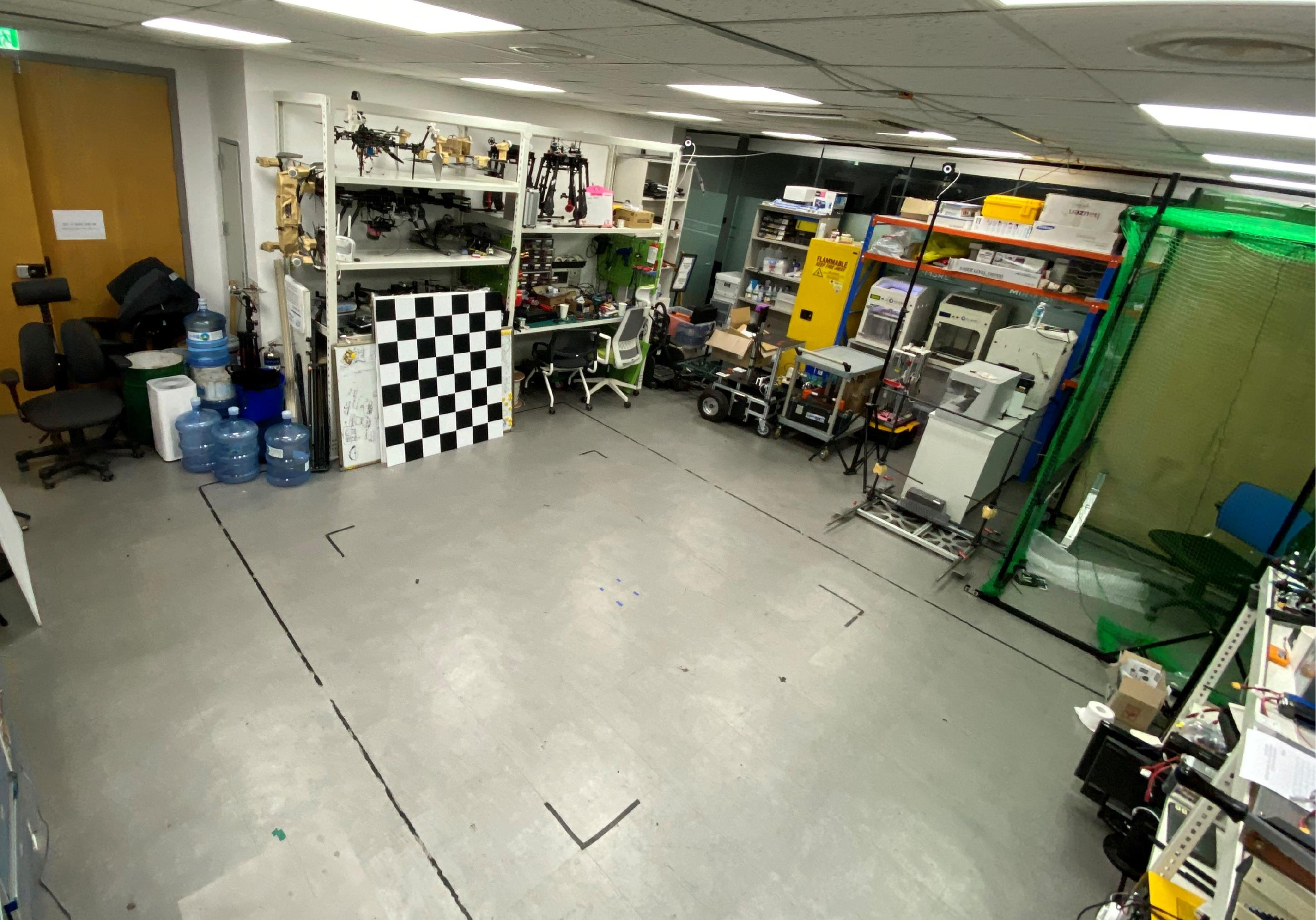}  
  \caption{}
  \label{fig:lab}
\end{subfigure}
\begin{subfigure}{0.5\textwidth}
    \centering
    \includegraphics[width=7.5cm]{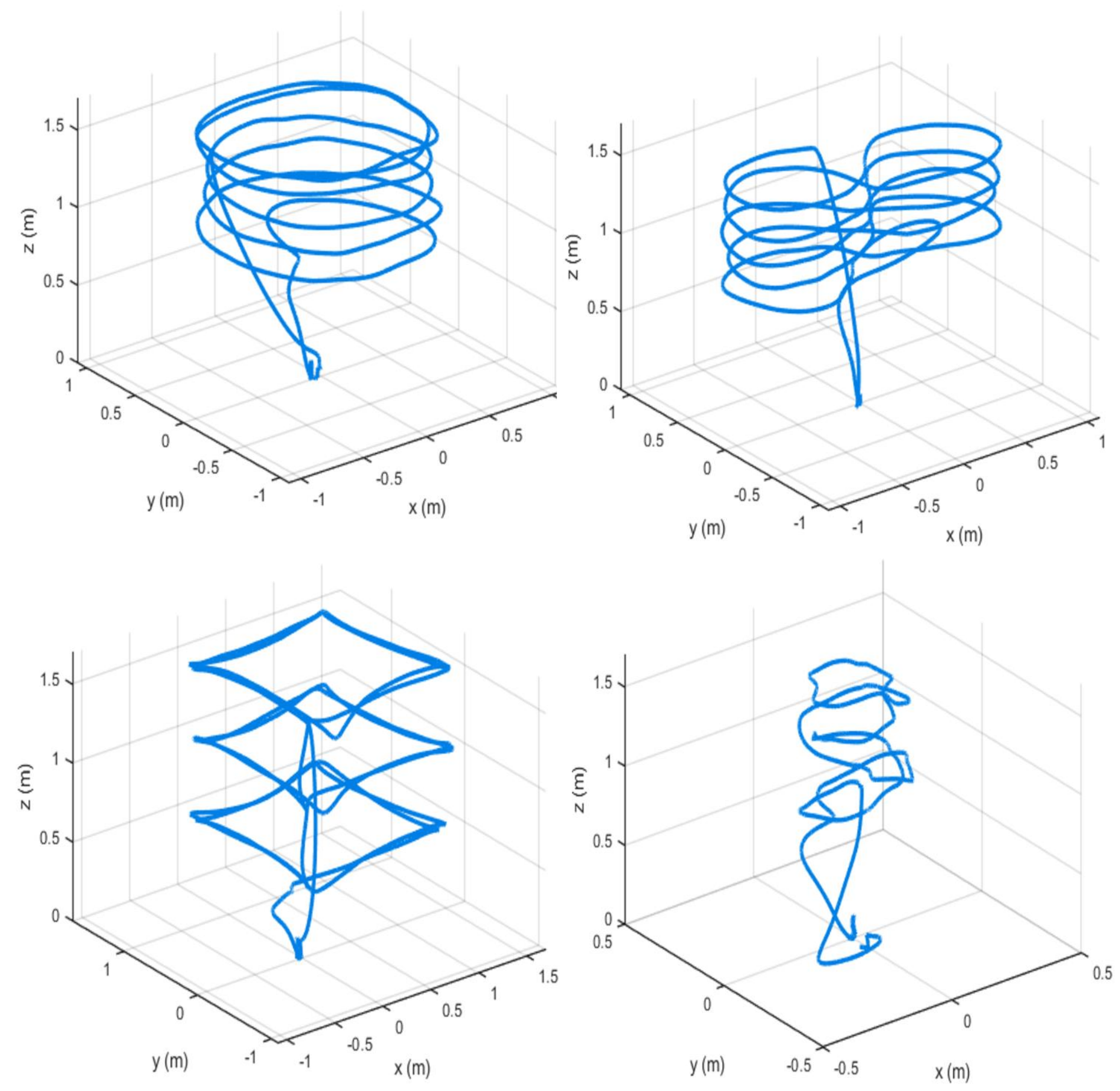}
    \caption{} 
    \label{fig:gt_total}
\end{subfigure}
 \caption{Experiment setup: (a) The UAV platform \textcircled{1} Intel Realsense D435i \textcircled{2} Pixhawk4 mini \textcircled{3} Jetson TX2 with a carrier board \textcircled{4} Reflective marker (b) Test environment (c) Ground Truth trajectories of KAIST VIO dataset}
\label{fig:fig1}
\vspace{-0.5cm}
\end{figure}


Visual-inertial odometry (VIO) algorithms are a representative example of the latter approach. A combination of inertial measurement units (IMUs) and the camera could solve the odometry problems more accurately and efficiently, complementing the {imperfections present in both technologies.} Numerous methodologies have been proposed for this combination, and several applications have been proposed\cite{kneip2011robust,li2013high},\cite{forster2014svo,leutenegger2015keyframe}.


{A recent trend is to combine deep learning with VO/VIO methods or use a GPU-accelerated front-end for those methods.} To achieve this, the hardware platform on which the algorithm runs should have sufficient resources. Therefore, NVIDIA Jetson boards equipped with graphics processing units (GPUs) {are used as they} have the potential to be used as a basic hardware platform in the future. Jetson boards are hardware modules released by NVIDIA and are developed to run software for autonomous machines. They are used as a companion computer for numerous autonomous robotic platforms, especially in UAVs, as they consume less power and overcome the limitations of the UAV platform in terms of size and weight. 

In addition, UAV applications that require real-time deep learning processes such as object detection and tracking as well as drone racing, use a lightweight network structure with Jetson boards installed. {To avoid the installation of} an additional embedded board for state estimation, the latest VIO algorithms should work well on these Jetson boards by sharing the computing resources with other processes. However, few studies have been conducted on the performance evaluation of VIO algorithms on various Jetson boards. 

This study aims to comprehensively analyze the feasibility and evaluate the performance of VIO algorithms, which are open {source}, and widely applied and used on various NVIDIA hardware configurations. We test three mono-VIO (VINS-Mono\cite{vins-mono}, ROVIO\cite{rovio-2}, and ALVIO\cite{alvio}), two stereo-VO (ORB-SLAM2 stereo\cite{orb2} and VINS-Fusion w/o IMU\cite{vins-fusion-1}), and four stereo-VIO (VINS-Fusion w/ IMU\cite{vins-fusion-1}, VINS-Fusion w/ GPU\cite{vinsfugpu}, Stereo-MSCKF\cite{msckf-vio}, and Kimera\cite{kimera}) algorithms and benchmark them on NVIDIA Jetson TX2, Xavier NX, and AGX Xavier boards, respectively. 

Furthermore, we conduct benchmark tests on the proposed dataset. The KAIST VIO dataset includes four different trajectories such as \texttt{circle}, \texttt{infinity}, \texttt{square}, and \texttt{pure\_rotation} with normal speed, high speed, and {head} rotation (Fig.~\ref{fig:fig1}(c)). Each sequence contains a pair of stereo images, one RGB image, and IMU data with accurate ground truth by a motion capture system acquired during UAV flight. {It is crucial to resolve the vulnerabilities caused by estimation error in visual-inertial state estimation that occurs during pure rotation\cite{olson2001stereo}.} {The dataset in this study consists of several rotation situations; hence, it is suitable to evaluate the performance or resistance encountered by each algorithm for hard cases to VIO.}

The main contributions of this study are as follows:


\begin{itemize}
\item {This study presents} the feasibility analysis and performance evaluation of various visual(-inertial) odometry algorithms on several NVIDIA Jetson boards, including the latest model "Xavier NX".
\item We propose a novel \textbf{KAIST VIO dataset} with different sets of sequences containing many rotations. The comparison shown in this paper presents an index of performance of the algorithm and Jetson board for motion trajectory with specific geometric and physical characteristics. The full dataset is available at: \url{https://github.com/zinuok/kaistviodataset}.
\end{itemize}

The rest of the paper is organized as follows: Section \ref{sec:related} reviews related works. Section \ref{sec:datasets} describes the proposed dataset. Section \ref{sec:exp} benchmarks the VIO algorithms with the dataset, and Section \ref{sec:result} analyzes the results in detail. Finally, Section \ref{sec:cons} summarizes our contributions and future works.



\section{Related works} \label{sec:related}

\subsection{Benchmark Comparison of VO/VIO}
Numerous studies {have been conducted} on the benchmarking of VO or VIO methods. Delmerico and Scaramuzza\cite{vio-benchmark} presented the overall benchmark comparisons of the state-of-the-art VIO algorithms on several hardware platforms (Laptop, Intel NUC, UP Board, and ORDROID) using the EuRoC dataset\cite{euroc}. However, the benchmark included only monocular visual-inertial methods, and not the stereo VO algorithms. Choi\cite{open-source-benchmark} presented a benchmark comparison of open-source methods based on \cite{euroc} and TUM VI dataset\cite{tum_vi}; {however, a comparison of various algorithms was absent.} 
Similarly, {the authors in \cite{monovo} presented the} benchmarking of vision-based odometry using their own dataset; however, only monocular VO methods were compared. For vision-based methods that require image processing tasks, an embedded system with a GPU might be an appropriate solution to accelerate the processing time. Giubilato \textit{et al.}\cite{vio-tx2-1} compared the well-known VO and SLAM methods on the Jetson TX2 platform.

\subsection{Benchmark Comparison of Jetson Boards}
There are several studies {that compared the} performance of Jetson boards. By using a deep-CNN algorithm, S{\"u}zen \textit{et al.}\cite{jetson-bench-1} compared Jetson TX2, Jetson Nano, and Raspberry Pi board with respect to accuracy and resource consumption. Ullah and Kim\cite{jetson-bench-2} presented the performance benchmarks of Jetson Nano, Jetson TX1, and Jetson AGX Xavier running deep learning algorithms that require complex computations, {in terms of resource consumption such as CPU, GPU, memory usage, and} processing time. Jo \textit{et al.}\cite{gpu-benchmark} also described a set of CNN benchmark comparisons of Jetson TX2, Jetson Nano, GTX1060, and Tesla V100. In existing studies, the performance comparison of the various Jetson boards targeting VO/VIO methods has not been clearly evaluated. Furthermore, the most recent Jetson NX board has not been included in the comparison.

\subsection{Benchmark Dataset in Harsh Environment}
{It is crucial to prove that} VO/VIO work well in a real environment with several harsh cases. Zu{\~n}iga-No{\"e}l \textit{et al.}\cite{corner-case-1} proposed an in/outdoor dataset in which low-texture scenes or scenes with dynamic illumination are included. These conditions are difficult cases of vision-based odometry. Kasper \textit{et al.}\cite{corner-case-2} also presented a dataset of scenes with dynamic motion blur, various degrees of illumination, and low camera exposure. Another study\cite{corner-case-3} analyzed the effect of photometric calibration, motion bias, and rolling shutter effect on the performance of vision-based methods. Pfrommer \textit{et al.}\cite{corner-case-4} introduced a dataset similar to \cite{corner-case-2},\cite{corner-case-3}. This includes partially rapid rotational motion; however, it is only a part of the entire path. It is still necessary to compare the performance of existing methods for the rotational movement itself in a rotation-only trajectory.





\section{DATASET} \label{sec:datasets}
{
The main contributions of KAIST VIO dataset are as follows:
\begin{itemize}
\item It includes pure-rotational and harsh motions for VIO that were not covered well in the other datasets.
\item Each trajectory sequence is subdivided into three types: normal/fast/head to ensure that benchmarking for each motion type is possible.
\end{itemize}
}

The data are recorded in the 3.15 $\times$ 3.60 $\times$ 2.50 m sized indoor laboratory as shown in Fig.~\ref{fig:fig1}(b). This environment has sufficient image features to run various VO/VIO algorithms. The KAIST VIO dataset provides four types of paths {with different geometrical properties.} To acquire accurate geometric characteristics of each trajectory, the drone (Fig.~\ref{fig:fig1}(a)) for data collection is automatically flown as programmed.

\subsection{Sensor Setup}\label{sec:sensor_setup}
The sensors and reference systems used for data collection are shown in Table \ref{table:sensor_setup}. Fig.~\ref{fig:fig1}(a) shows the camera and IMU mounted on the drone body.
\vspace{-0.2cm}

\begin{table}[t]
\renewcommand{\arraystretch}{0.7}
\renewcommand{\tabcolsep}{0.67mm}
\caption{Sensor setup}
\label{table:sensor_setup}
\begin{center}
\begin{tabular}{c|c|c|c}
\hline
Sensor & Type & Data & Rate\\
\hline
\multirow{2}{*}{Camera} & \multirow{2}{*}{D435i} & IR 1,2 (640$\times$480) & 30 Hz\\
&  & RGB (640$\times$480) & 30 Hz\\
\multirow{2}{*}{IMU} & \multirow{2}{*}{Pixhawk 4 mini} & 3-axes accel., & 100 Hz\\
& & 3-axes gyro.& 100 Hz\\
Ground Truth & OptiTrack Mocap & Ground Truth & 50 Hz\\
\hline
\end{tabular}
\end{center}
\vspace{-0.7cm}
\end{table}


\noindent
\newline\textbf{Camera } Images with 640 $\times$ 480 resolution are obtained by Intel Realsense D435i, mounted in front of the drone {to ensure that it can look forward} at a rate of 30 Hz. The rolling shutter and global shutter deliver RGB images and infra-red (IR) images with the emitter turned off, respectively. In this study, for benchmarking the VO/VIO algorithms, only the IR images were used {as the global shutter is more suitable for rapid motion in this dataset.}

\noindent
\textbf{IMU } IMU data are logged at a rate of 100 Hz using Pixhawk4 mini, mounted at the center of the drone. A VI sensor unit consists of this IMU and D435i camera. {Kalibr\cite{kalibr}
is used to obtain spatial and temporal calibration data. To accomplish this, the VI sensor unit records a unique pattern (AprilTag) with smooth  6-DOF motions. Kalibr uses a temporal basis function to calculate the time offset between the camera and IMU. In addition, temporal synchronization is performed using high-rate IMU data accumulated and interpolated for each camera frame.  
}
Furthermore, the noise parameter values of the Pixhawk4 mini are calculated using a Kalibr$\_$allan\cite{kalibr_allan}. This allows more accurate calibration data to be obtained {in addition to} optimal parameter tuning for VO/VIO algorithms.

\noindent
\textbf{Ground truth } To obtain the accurate ground truth, an OptiTrack Prime\textsuperscript{X} 13 motion capture system\cite{mocap} consisting of six cameras is used. {This motion capture system captures 6-DOF motion information by tracking the motion capture marker mounted on top of the drone.} The information is recorded at a rate of 50 Hz within millimeter accuracy during the flight. Additionally, a transformation matrix for aligning the difference in positions between the origin of the ground truth defined by five markers and the VI sensor unit is included in the dataset format.

\subsection{Dataset Format}
\begin{figure}[h]
    \centering
    \includegraphics[width=62mm, height=60mm]{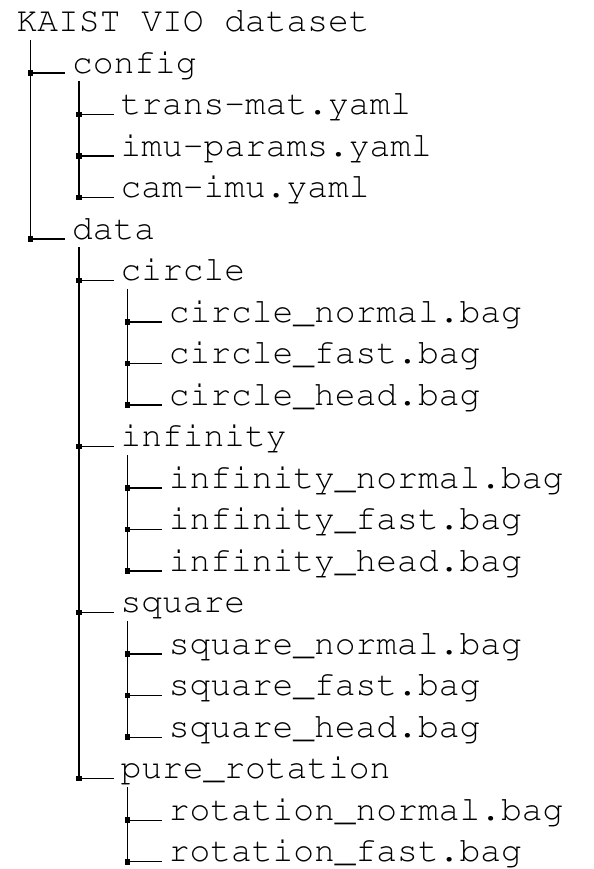}


\caption{KAIST VIO dataset structure }
\label{fig:tree}
\vspace{-0.3cm}
\end{figure}

This dataset has two sub-directories, the \texttt{config} and \texttt{data} directories, as shown in Fig. \ref{fig:tree}.
\vspace{-0.15cm}


\noindent
\textbf{\texttt{config} directory }  The config directory contains three YAML files. \texttt{trans-mat.yaml} contains translational matrix information for correcting the offset as described in Section \ref{sec:datasets}.\textit{A}. This offset has already been applied to the ground truth of the Robot Operating System (ROS) bag data but has been included for 
reference. \texttt{imu-params.yaml} contains four noise parameter estimates for Pixhawk4 mini: {white noise of the gyroscope, white noise of the accelerometer, random walk of the gyroscope, and random walk of the accelerometer.} These values are obtained by based on \cite{kalibr_allan}. \texttt{cam-imu.yaml} contains the calibrated data from the VI sensor unit.

\noindent
\textbf{\texttt{data} directory } Each set of data is recorded as a bag file, a file format commonly used in ROS. Each file stores the sensor information required to run the algorithms acquired from the camera and the IMU. Additionally, the ground truth 6-DOF pose information of the drone, acquired using the motion capture system, is saved. All the data in each file are recorded in the form of ROS topics during flight. There are a total of four sub-directories with different geometric classifications of the motion trajectories: \texttt{circle}, \texttt{infinity}, \texttt{square}, and \texttt{pure\_rotation} (see Fig. 1(c)). Furthermore, each sub-directory contains several types of data: \texttt{normal} (normal speed with fixed heading), \texttt{fast} (high speed with fixed heading), and \texttt{{head}} (normal speed with rotational motion). For details, {please refer to our dataset link.}






\begin{table}[!t]
\centering
\renewcommand{\arraystretch}{0.9}
\renewcommand{\tabcolsep}{0.5mm}
\caption{Specifications of Jetson platforms}
{\scriptsize
\begin{center}

\begin{tabular}{p{1cm}|p{2.3cm}|p{2.3cm}|p{2.4cm}}
\hline
& \hfil Jetson TX2 & \hfil Xavier NX & \hfil AGX Xavier\\
\hline
\hfil CPU & 6-core Denver and A57 & \hfil 6-core Carmel ARM & \hfil 8-Core Carmel ARM\\
\hfil GPU & \hfil 256 Core Pascal & \hfil 384 Core Volta & \hfil 512 Core Volta\\
Memory & \hfil 8GB 128bit LPDDR4 & \hfil 8GB 128bit LPDDR4x & 32GB 256bit LPDDR4x\\
\hfil Size(mm) & \hfil 50$\times$110$\times$37 & \hfil 100$\times$90$\times$32 & \hfil 105$\times$105$\times$65\\
\hfil Weight & \hfil 211 g (with J120\cite{j120}) & \hfil 184.5 g  & \hfil 670 g\\
\hfil Power & \hfil 7.5W(or 15W) & \hfil 10W(or 15W, 30W) & \hfil 10W(or 15W, 30W)\\
\hline
\end{tabular}
\end{center}

}
\label{table:jetson_platform}
\vspace{-0.5cm}
\end{table}

\section{EXPERIMENTS}\label{sec:exp}
\subsection{Compared Hardware Platforms}\label{sec:jetson}

NVIDIA Jetson boards are used as a hardware platform for performance comparison. The Jetson platforms used in this study are Jetson TX2, Jetson AGX Xavier, and the recently released Jetson Xavier NX. A brief description of each board is as follows, and the detailed specification for each platform is shown in Table \ref{table:jetson_platform}:
\begin{itemize}
\item Jetson TX2: TX2 is widely used as a companion computer for UAV systems {owing to its better CPU and GPU performance as well as larger memory than that of} Nano and TX1.
\item Jetson Xavier NX: Xavier NX is a module recently released by NVIDIA. Owing to its small size and low weight similar to the Jetson Nano, it is suitable for robotic systems having significant physical limitations.
\item Jetson AGX Xavier: AGX Xavier has a decent performance and can serve as a workstation for the autonomous system. It is mainly used for industrial robots and large UAV systems.
\end{itemize}

\subsection{Compared Algorithms}
Table \ref{table:algorithm} shows all algorithms compared in this paper.
\vspace{-0.1cm}
\begin{table}[h]
\renewcommand{\arraystretch}{0.9}
\caption{Compared algorithms: monocular and stereo}
\begin{center}
{\scriptsize
\begin{tabular}{c|c c}
\hline
Monocular & \multicolumn{2}{c}{Stereo}\\
w/ IMU & w/o IMU & w/ IMU\\
\hline
VINS-Mono\cite{vins-mono} & VINS-Fusion\cite{vins-fusion-1} & VINS-Fusion-gpu\cite{vinsfugpu}\\
ALVIO\cite{alvio} & ORB-SLAM2\cite{orb2} & VINS-Fusion-imu\cite{vins-fusion-1}\\
ROVIO\cite{rovio-2} &  & Stereo-MSCKF\cite{msckf-vio}\\
 &  & Kimera\cite{kimera}\\
\hline
\end{tabular}
}
\end{center}
\label{table:algorithm}
\vspace{-0.55cm}
\end{table}


\subsection{Evaluation}
All Jetson boards are set to the maximum CPU clock mode, consuming maximum power to thoroughly compare the potential {performance of each algorithm.}
The performance evaluation is performed based on the resource usage and {Absolute} Trajectory Error (ATE) for each algorithm and platform.
Considering the usage of resources such as CPU, memory, and GPU, all algorithms are measured in \texttt{infinity\_fast} path, where all algorithms do not diverge and have sufficient dynamic motion. To obtain more accurate measurements, only necessary processes are run, which is {intended by the authors of each algorithm to obtain the appropriate trajectory estimation.} The total sum of the resource usage is recorded every 0.1 seconds.
For calculating the ATE, the origin alignment method \cite{grupp2017evo} is used for aligning the ground truth and estimated odometry values.

\subsubsection{Setup}
Ubuntu 18.04 with ROS melodic was setup on all platforms. Jetpack 4.2 was installed on TX2 and 4.4 on AGX Xavier and Xavier NX. Benchmark evaluation uses the data sequences described in Section \ref{sec:datasets}.

\subsubsection{Parameter setting}

For each algorithm, the trade-off between resource usage and accuracy is considerably different, and this study aims to present a performance comparison considering both of them for a general UAV system. Therefore, considering the trade-off, parameter values are tuned, maintaining widely used preset values recommended by the author(s) of each algorithm. The parameter settings of the used algorithms are as follows:

\noindent
\textbf{VINS-Mono}: 
The maximum number of features is set to 150, and the minimum distance between the two features is set to 25. The loop closure is disabled.

\noindent
\textbf{ALVIO}: 
ALVIO (adaptive line visual-inertial odometry) is an algorithm that additionally introduces line features to the existing VINS-Mono and overcomes the failure of line tracking through an optical-flow-based method. The parameter setting is similar to VINS-Mono.

\noindent
\textbf{ROVIO}: 
The maximum number of features is set to 20 and the size of the patch is set to 6 pixels. The maximum distance penalized during the bucketing process is set to 100 pixels.

\noindent
\textbf{ORB-SLAM2 stereo}: 
The maximum number of features per frame is set to 1200.

\noindent
\textbf{VINS-Fusion}: 
The maximum number of features is set to 350 and the minimum distance between the two features is set to 30. {The IMU and loop closure are disabled.}

\noindent
\textbf{VINS-Fusion-gpu}: 
Same as VINS-Fusion except that the GPU and IMU are enabled.

\noindent
\textbf{VINS-Fusion-imu}: 
Same as VINS-Fusion except that the IMU is enabled.

\noindent
\textbf{Stereo-MSCKF}: 
The minimum and maximum number of features per grid (3 $\times$ 4 grid that divides one image frame) constituting a frame are set to 3 and 4, respectively. The patch size is set to 15.

\noindent
\textbf{Kimera}: 
The maximum number of features is set to 800, and the minimum distance between the two features is set to 8. The IMU pre-integration type is a non-combined IMU factor method.

\section{RESULTS ANALYSIS}\label{sec:result}
\begin{figure*}[t]
    \centering
    \includegraphics[width=17.5cm, height=11cm]{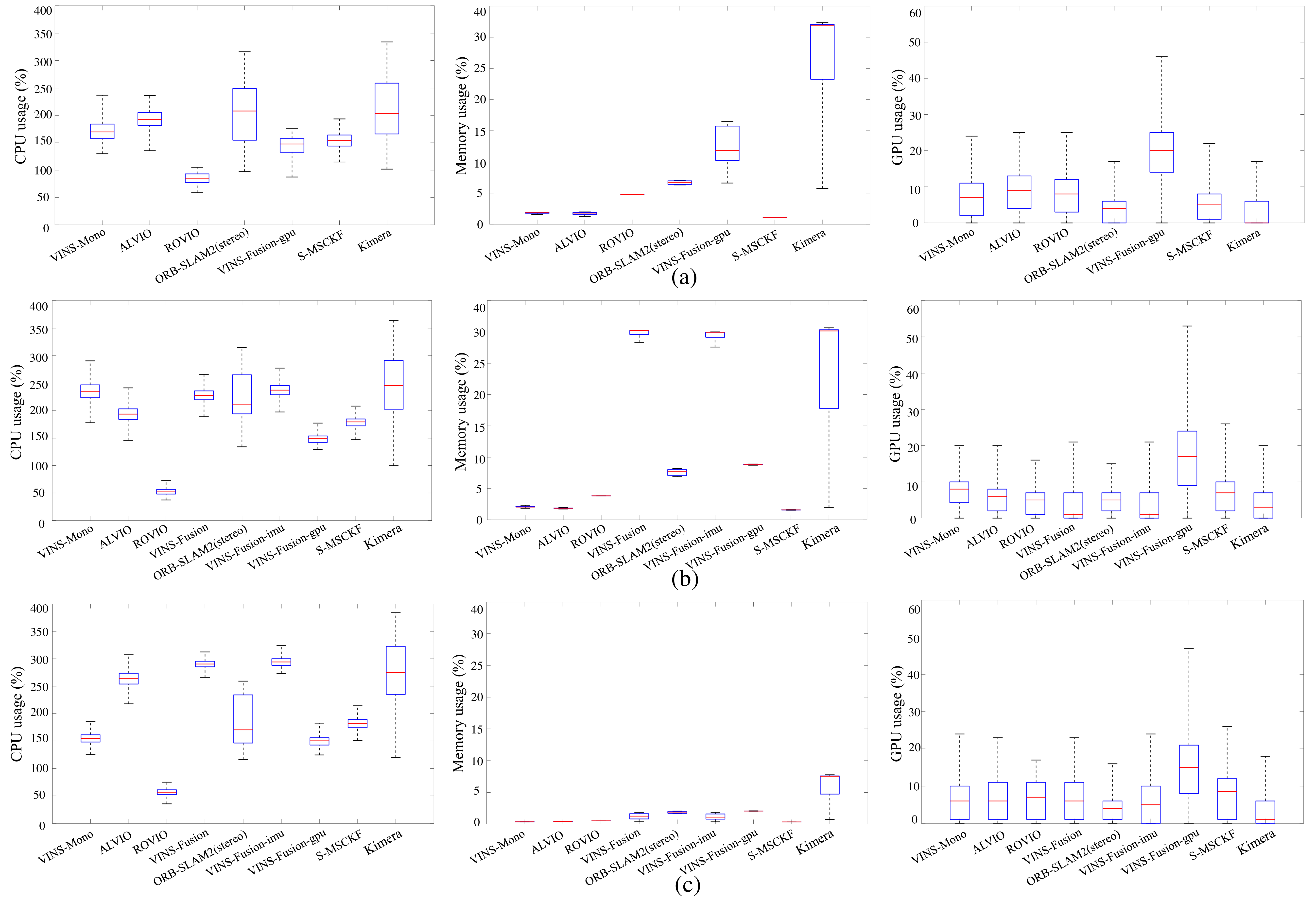}
    \caption{Statistical comparison of CPU, memory, GPU usages  for possible algorithm-platform combinations on {\texttt{infinity\_fast} sequence} (a) TX2 (VINS-Fusion and VINS-Fusion-imu fail on all sequences) (b) NX (c) Xavier} 
    \label{fig:resource}
    \vspace{-0.4cm}
\end{figure*}
\subsection{Analysis of Resource Usage}
The CPU, memory, and GPU usages are shown in Fig. \ref{fig:resource}. On TX2, the GPU-accelerated version of VINS-Fusion (VINS-Fusion-gpu) was used because VINS-Fusion and VINS-Fusion-imu do not run owing to insufficient memory and CPU performance issues.

Considering the CPU usage, Kimera and ORB-SLAM2 stereo were loosely-bounded and had relatively higher values than those of other algorithms on all Jetson platforms. {They needed a larger number of features per frame compared with those of} other algorithms, and the variation in CPU usage was considerable, depending on the number of detected features (0 to 800 and 0 to 1200 in this case). The CPU usage of ROVIO was the lowest on all Jetson platforms as ROVIO tracks the patch extracted from the detected feature and reduces the computation compared with that of other algorithms. {Except for ROVIO, all other algorithms showed more than 100\% CPU usage on each platform because of multi-core processing.} There was no significant difference between the mono and stereo algorithms.

{The memory usage of} stereo VO/VIO algorithm was higher than that of the monocular VIO algorithm on all Jetson platforms. The memory usage of Stereo-MSCKF was similar to that of the monocular-based algorithm as the number of used features per frame is small (3 or 4 features per grid) and it is a filtering-based method.
Furthermore, among the stereo-based methods, VINS-Fusion and VINS-Fusion-imu showed higher usage rates than other algorithms except Kimera. This tendency was relatively significant on Xavier NX, which has a lower CPU performance than AGX Xavier. The memory usage of Kimera was considerably higher than that of other algorithms on all Jetson platforms as Kimera requires numerous computations per keyframe. For TX2, which lacks CPU performance, this difference was more noticeable. When comparing each Jetson platform, AGX Xavier has significantly lower memory usage than the rest as it has the most massive memory of 32 GB.  

{VINS-Fusion-gpu had the highest GPU usage} on all Jetson platforms because it is the only algorithm that uses GPU-acceleration. The GPU usage by the Jetson platform did not show any significant difference. The same was true for the GPU usage of stereo and monocular-based systems in each platform.
Considering the overall result, these three platforms are sufficient for the algorithms that use GPU-acceleration without any constraints.

\subsection{Analysis of ATE RMSE}
\begin{table*}[t]
\caption{RMSE (Unit: m) of Absolute Trajectory Error (ATE) for all data sequences. We aligned the estimated trajectory to the ground truth trajectory according to the origin alignment method. The best performance combinations in each sequence on each platform are highlighted in \textbf{bold}. `\ding{53}' denotes the diverged one and `-' denotes failed to run, respectively. \newline\small{(cir: \texttt{circle}, inf: \texttt{infinity}, squ: \texttt{square}, rot: \texttt{rotation}, and n: \texttt{normal}, f: \texttt{fast}, h: \texttt{head})}}
\centering
\renewcommand{\arraystretch}{0.85}
\renewcommand{\tabcolsep}{0.9mm}
{\scriptsize
\begin{tabular}{c|ccccccccc|ccccccccc|ccccccccc}
\hline
 & \multicolumn{9}{c}{\textbf{\tiny{TX2}}} & \multicolumn{9}{c}{\textbf{\tiny{NX}}} & \multicolumn{9}{c}{\textbf{\tiny{Xavier}}}\\
 & \rotatebox[origin=c]{90}{\tiny{VINS-Mono}} &\rotatebox[origin=c]{90}{\tiny{ALVIO}} &\rotatebox[origin=c]{90}{\tiny{ROVIO}} &\rotatebox[origin=c]{90}{\tiny{Kimera}} &\rotatebox[origin=c]{90}{\tiny{VINS-Fusion}} &\rotatebox[origin=c]{90}{\tiny{VINS-Fusion-gpu}} &\rotatebox[origin=c]{90}{\tiny{VINS-Fusion-imu}} &\rotatebox[origin=c]{90}{\tiny{ORB-SLAM2}} &\rotatebox[origin=c]{90}{\tiny{S-MSCKF}}
 & \rotatebox[origin=c]{90}{\tiny{VINS-Mono}} &\rotatebox[origin=c]{90}{\tiny{ALVIO}} &\rotatebox[origin=c]{90}{\tiny{ROVIO}} &\rotatebox[origin=c]{90}{\tiny{Kimera}} &\rotatebox[origin=c]{90}{\tiny{VINS-Fusion}} &\rotatebox[origin=c]{90}{\tiny{VINS-Fusion-gpu}} &\rotatebox[origin=c]{90}{\tiny{VINS-Fusion-imu}} &\rotatebox[origin=c]{90}{\tiny{ORB-SLAM2}} &\rotatebox[origin=c]{90}{\tiny{S-MSCKF}}
 & \rotatebox[origin=c]{90}{\tiny{VINS-Mono}} &\rotatebox[origin=c]{90}{\tiny{ALVIO}} &\rotatebox[origin=c]{90}{\tiny{ROVIO}} &\rotatebox[origin=c]{90}{\tiny{Kimera}} &\rotatebox[origin=c]{90}{\tiny{VINS-Fusion}} &\rotatebox[origin=c]{90}{\tiny{VINS-Fusion-gpu}} &\rotatebox[origin=c]{90}{\tiny{VINS-Fusion-imu}} &\rotatebox[origin=c]{90}{\tiny{ORB-SLAM2}} &\rotatebox[origin=c]{90}{\tiny{S-MSCKF}}\\
\hline
cir-n & 0.10 &\textbf{0.07} &\ding{53} &0.08 &- &0.08 &- &0.10 &0.14 &0.13 &0.09 &\ding{53} &0.12 &\textbf{0.06} &0.09 &0.11 &0.09 &0.12 &0.12 &0.12 &\ding{53} &0.08 &\textbf{0.07} &0.09 &0.08 &0.08 &0.11\\
cir-f &\textbf{0.07} &0.13 &0.79 &0.13 &- &0.14 &- &0.28 &0.12 &0.15 &\textbf{0.05} &0.83 &0.07 &0.12 &0.13 &0.10 &0.11 &0.19 &0.14 &0.12 &0.80 &\textbf{0.08} &0.16 &0.13 &0.13 &0.12 &0.23\\
cir-h &0.24 &\ding{53} &2.11 &0.25 &- &\textbf{0.10} &- &0.19 &\textbf{0.10} &0.43 &0.45 &2.12 &0.28 &\textbf{0.08} &0.11 &0.13 &0.13 &0.21 &0.41 &0.49 &2.11 &0.26 &\textbf{0.06} &0.11 &0.07 &0.15 &0.20\\
\hdashline
inf-n & 0.14 &0.11 &\ding{53} &\textbf{0.09} &- &\textbf{0.09} &- &0.35 &0.10 &0.10 &0.12 &1.32 &\textbf{0.05} &\textbf{0.05} &0.09 &0.08 &0.08 &0.32 &0.24 &0.12 &1.19 &0.09 &\textbf{0.07} &0.09 &\textbf{0.07} &\textbf{0.07} &0.09\\
inf-f & 0.11 &\textbf{0.06} &0.44 &{0.19} &- &\textbf{0.06} &- &0.22 &0.20 &0.08 &0.07 &0.41 &0.14 &0.09 &\textbf{0.05} &{0.08} &0.10 &0.17 &0.10 &0.09 &0.44 &{0.13} &{0.07} &\textbf{0.05} &{0.07} &0.07 &0.12\\
inf-h & 1.19 &\ding{53} &\ding{53} &0.94 &- &\textbf{0.13} &- &1.50 &1.11  &0.50 &1.10 &\ding{53} &1.08 &\textbf{0.12} &0.14 &\textbf{0.12} &\textbf{0.12} &0.60 &0.57 &0.48 &\ding{53} &1.09 &\textbf{0.09} &0.14 &0.12 &\textbf{0.09} &0.87\\
\hdashline
squ-n & 0.20 &0.16 &0.46 &\textbf{0.08} &- &0.12 &- &0.44 &0.18 &0.17 &0.16 &0.46 &0.17 &0.17 &0.12 &0.21 &\textbf{0.09} &0.10 &0.11 &\textbf{0.10} &0.47 &0.13 &0.17 &\textbf{0.10} &0.15 &0.12 &0.15\\
squ-f &\textbf{0.07} &0.13 &0.56 &0.14 &- &0.10 &- &0.29 &0.17 &0.14 &\textbf{0.07} &0.56 &0.19 &\textbf{0.07} &0.11 &0.13 &0.09 &0.30 &0.12 &0.10 &0.56 &0.14 &\textbf{0.08} &0.11 &0.10 &0.14 &0.17\\
squ-h & \ding{53} &\ding{53} &\ding{53} &0.18 &- &\textbf{0.15} &- &0.16 &0.40 &0.34 &\ding{53} &\ding{53} &1.57 &0.19 &\textbf{0.15} &0.20 &0.16 &0.30 &0.30 &0.36 &\ding{53} &1.50 &0.18 &\textbf{0.15} &0.18 &0.17 &0.50\\
\hdashline
rot-n & 0.83 &\ding{53} &\ding{53} &0.16 &- &\textbf{0.12} &- &0.31 &0.16 &\ding{53} &\ding{53} &\ding{53} &0.17 &0.11 &0.12 &0.16 &0.17 &\textbf{0.10} &\ding{53} &0.81 &\ding{53} &0.18 &0.11 &0.12 &0.11 &0.16 &\textbf{0.07}\\
rot-f & \ding{53} &\ding{53} &2.74 &0.85 &- &\textbf{0.11} &- &0.18 &0.29 &0.40 &\ding{53} &\ding{53} &0.74 &0.28 &0.11 &\textbf{0.10} &0.21 &0.29 &0.89 &0.72 &\ding{53} &0.90 &0.26 &0.11 &\textbf{0.07} &0.18 &0.19\\
\hline
\end{tabular}
}
\label{table:overall_rmse}
\end{table*}

\begin{figure*}[ht]
    \vspace{-0.25cm}
    \centering
    \includegraphics[width=17cm, height=10cm]{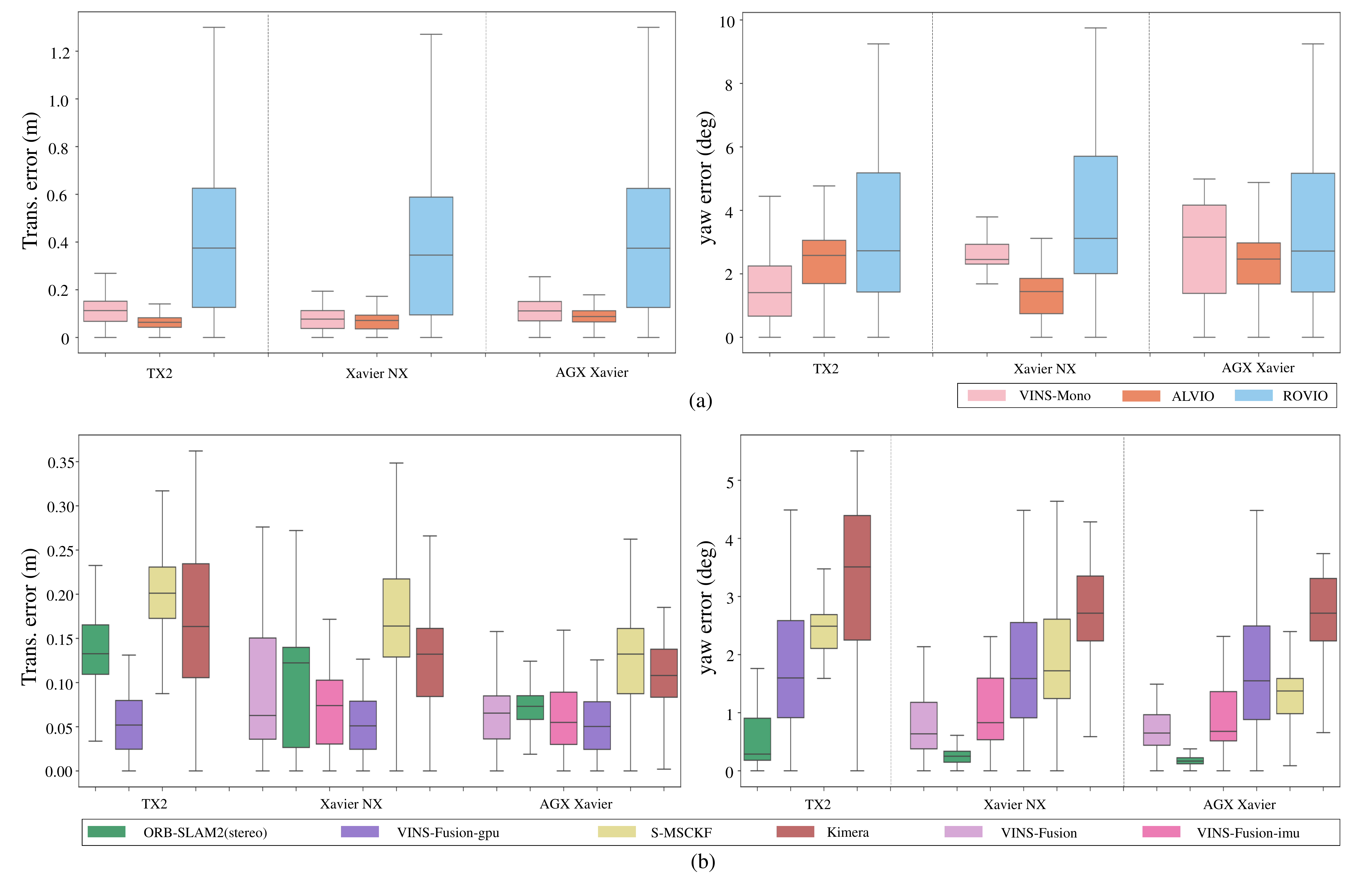}
    \caption{Boxplot for translational and yaw errors on {\texttt{infinity\_fast} sequence.} RMSE errors were calculated using \cite{grupp2017evo} (a) Monocular VIO (b) Stereo VO/VIO} 
    \label{fig:error}
    \vspace{-0.7cm}
\end{figure*} 

\begin{figure*}[ht]
    \centering
    \includegraphics[width=17.5cm]{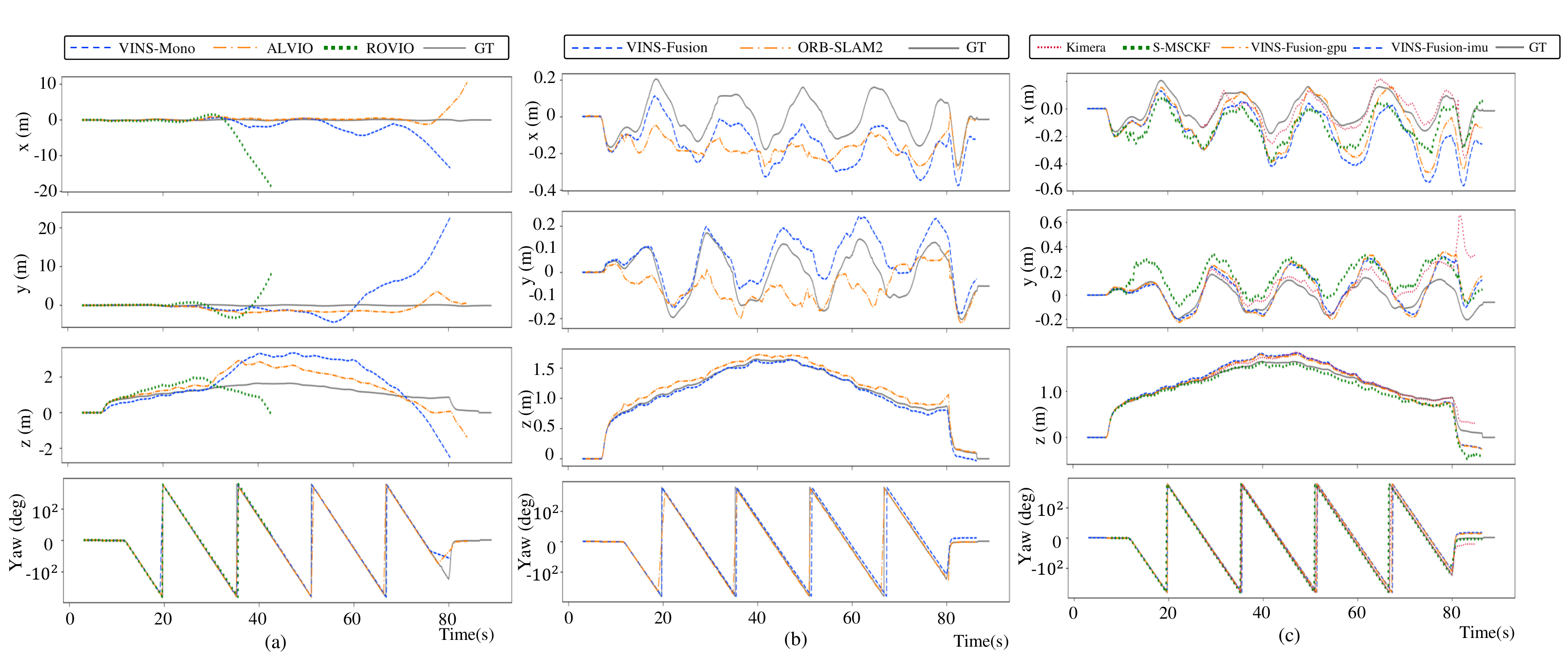}
    \caption{{Resulting trajectories} of VI/VIO tests on \texttt{rotation\_normal} sequence running on Jetson NX board. We aligned the estimated trajectory
to the ground truth trajetory according to the origin alignment method. (a) Monocular VIO (b) Stereo VO (c) Sterero VIO} 
    \label{fig:rotation}
    \vspace{-0.6cm}
\end{figure*}
The ATE RMSE (RMSE of Absolute Trajectory Error) for all trajectories, Jetson boards, and algorithms are shown in Table~\ref{table:overall_rmse}. On each platform, the algorithm that exhibits the smallest error for each trajectory sequence is highlighted in \textbf{bold}. All 11 sequences were recorded in the same environment.
Therefore, there is a difference in the feature displacement of two consecutive frames for each path, and it is necessary to analyze the error by considering the {motion characteristics of each path.}

In the KAIST VIO dataset, a representative sequence with no rotational motion and only rapid translational motion is \texttt{infinity\_fast}. Translational and yaw errors for each algorithm and platform for this sequence are shown in Fig. \ref{fig:error}. For translational errors, stereo methods generally performed better than monocular-based methods on all platforms. VINS-Mono and ALVIO showed excellent performance similar to stereos, and ALVIO was better than VINS-Mono. Adding a line-feature (multi-pixels) to a point-feature (single-pixel), ALVIO could precisely track the extracted features without losing them. This robustness was also shown for yaw error except for TX2, which has an unsatisfactory performance to run ALVIO. \texttt{rotation\_normal} and \texttt{rotation\_fast} sequences, which have little translational movement, provide harsh paths for VO/VIO. Hence, the estimated odometry value often diverged in many algorithms. For the \texttt{rotation\_normal} sequence, the x, y, z, and yaw errors of each algorithm executed on Xavier NX are shown in Fig. \ref{fig:rotation}. The first algorithm diverged was ROVIO, and ROVIO mostly diverged in sequences with rotational motion: \texttt{infinity\_head}, \texttt{square\_head}, \texttt{rotation\_normal}, and \texttt{rotation\_fast}. ROVIO showed weak rotational motion because multi-level patches are not properly extracted or tracked during rapid scene transitions.

The overall result showed robustness against rotations in the order of stereo VIO, stereo VO, and mono VIO. However, all three methods showed excellent performance {for yaw errors.}
Comparing VINS-Fusion, VINS-Fusion-imu, and VINS-Fusion-gpu in the \texttt{rotation} sequences, the following two tendencies were observed. In \texttt{rotation\_normal}, VINS-Fusion showed a smaller error than VINS-Fusion-imu and VINS-Fusion-gpu. In \texttt{rotation\_fast}, the error of VINS-Fusion-imu and VINS-Fusion-gpu was smaller than that of VINS-Fusion (see Table \ref{table:overall_rmse}). This is because the IMU is specialized in detecting rapid motion, and the camera is specialized in detecting relatively slow motion. Moreover, the VINS-Fusion series {is considerably affected by the IMU as} IMU measurements are locally integrated with their pre-integration model, and their estimator refines extrinsic parameters between the camera and IMU online at the start of the flight. Therefore, for rotational motion, the VINS-Fusion series requires precise tuning of IMU parameters.

{Although the same algorithm with the fixed-parameter setting was run in the same sequence, the error on each board was different. The statistical characteristics of these differences are shown in Fig. \ref{fig:error}. For mono methods, ROVIO did not show any significant difference among boards in both translation/yaw errors. This implies that each board has sufficient computing resources to run ROVIO smoothly. Similarly, for VINS-Mono and ALVIO, no significant difference was observed among the boards in a translation error. For stereo methods, VINS-Fusion-gpu, which mainly depends on GPU operation, did not show any significant difference among the boards in both translation/yaw errors. This means that the GPU resources of each board are sufficient to run the VINS-Fusion-gpu smoothly. In ORB-SLAM2(stereo), S-MSCKF, and Kimera, translation/yaw errors were the highest in the order of TX2, Xavier NX, and AGX Xavier. This is because these algorithms were particularly limited by the computational performance of the board, and the computational performances of TX2 and Xavier NX were inferior to that of AGX Xavier to run these algorithms smoothly. This was consistent with the differences in the number of cores and the performance of CPU/GPU mounted on each board, as shown in Table \ref{table:jetson_platform}. Similarly, in VINS-Fusion and VINS-Fusion-imu, the translation/yaw error range was higher in Xavier NX than in AGX Xavier.} 

On the TX2 platform, VINS-Fusion-gpu showed the best performance for the trajectories with rotational motion. This is because VINS-Fusion-gpu is the only algorithm that uses the GPU to compensate for the insufficient computational performance of the CPU of TX2. Stereo methods, which perform computations using only the CPU without the GPU, have a larger error than monocular-based methods owing to the limitation of per-frame processing time. Compared with other platforms, the monocular-based algorithms had better performance than that of the stereo algorithms in TX2, except for cases that diverge on trajectories with rotational motion. On NX and Xavier, which have better CPU and memory performance than TX2, stereo methods were better than monocular-based ones. The overall error for each path was lower in Xavier than in NX. This is because Xavier has a better CPU and memory than NX, and the per-frame processing time is shorter than NX.
\section{CONCLUSIONS} \label{sec:cons}




This study presented a novel KAIST VIO dataset that has harsh trajectories for VO/VIO, and the overall performance of various VO/VIO algorithms (mono, stereo, and stereo + IMU) was evaluated on NVIDIA Jetson TX2, Xavier NX, and AGX Xavier platforms. {The goal of this study} was to benchmark well-known VO/VIO algorithms {using the proposed} dataset, which has considerable rotational movement, with hardware that has limited computing power{, is compact,} and has GPU cores.  

In summary, {the monocular VO/VIO would be suitable for use} in TX2. In stereo VO/VIO, a GPU-accelerated algorithm would be appropriate for use in TX2. For the UAV system, Xavier NX will be appropriate, given that the UAV system has physical limitations (payload, dimensions etc.). {In the absense of limitations,} AGX Xavier would be a better choice. In the rotational motion case, the stereo VIO method is robust for rapid rotation and the stereo VO is suitable for relatively slow rotation. 
The error in pure rotation movement is a huge challenge that VO/VIO must overcome. Therefore, this KAIST VIO dataset includes various pure rotational trajectories to serve as a benchmark tester to solve this problem. 

These results and the dataset presented in this paper can be used as an index for {determining the suitable pair of platform and algorithm for the UAV systems that fly along predefined} paths with certain motion characteristics. Please refer to our official link that has the descriptions of our dataset and the setting instructions on how to run each algorithm on Jetson boards. 

Run your VO/VIO algorithms on NVIDIA Jetson boards with our dataset to demonstrate its robustness for rotational motion.

\bibliographystyle{IEEEtran}
\bibliography{references}

\end{document}